\documentclass[runningheads]{llncs}

\usepackage{amsmath,graphicx,stfloats}
\usepackage[table,xcdraw]{xcolor}
\usepackage{graphicx}
\usepackage{cite}
\usepackage{hyperref}
\hypersetup{
    colorlinks=true,
    linkcolor=blue,
    filecolor=magenta,      
    urlcolor=cyan,
    pdftitle={Overleaf Example},
    pdfpagemode=FullScreen,
    }
\newcommand{\repeatthanks}{\textsuperscript{\thefootnote}}
\begin{document}
\title{MaskMTL: Attribute prediction in masked facial images with deep multitask learning}
\titlerunning{Attribute prediction}
%
\author{Prerana Mukherjee \inst{1} \thanks{Equal Contribution}\and
Vinay Kaushik  \inst{2} \repeatthanks\and
Ronak Gupta \inst{2} \repeatthanks \and
Ritika Jha  \inst{1} \repeatthanks \and
Daneshwari Kankanwadi  \inst{1} \repeatthanks \and
Brejesh Lall \inst{2} \repeatthanks
}
\authorrunning{Prerana et.al.}
\institute{School of Engineering, Jawaharlal Nehru University\\
\and
Department of Electrical Engineering, Indian Institute of Technology Delhi\\
}

\maketitle              
\begin{abstract}

Predicting attributes in the landmark free facial images is itself a challenging task which gets further complicated when the face gets occluded due to the usage of masks. Smart access control gates which utilize identity verification or the secure login to personal electronic gadgets may utilize face as a biometric trait. Particularly, the Covid-19 pandemic increasingly validates the essentiality of hygienic and contactless identity verification. In such cases, the usage of masks become more inevitable and performing attribute prediction helps in segregating the target vulnerable groups from community spread or ensuring social distancing for them in a collaborative environment. We create a masked face dataset by efficiently overlaying masks of different shape, size and textures to effectively model variability generated by wearing mask. This paper presents a deep Multi-Task Learning (MTL) approach to jointly estimate various heterogeneous attributes from a single masked facial image. Experimental results on benchmark face attribute UTKFace dataset demonstrate that the proposed approach supersedes in performance to other competing techniques.
The source code is available at 
\href{https://github.com/ritikajha/Attribute-prediction-in-masked-facial-images-with-deep-multitask-learning}{here.}

\end{abstract}
\section{Introduction}
\label{sec:intro}
With the ongoing COVID-19 pandemic situation, it is imperative to provide contactless and unhindered running operations, especially in public spaces like airport, railways, offices, tourist locations etc. Facial biometric based analysis can be an important contactless trait. Wearing masks to avert the spread of communicable diseases has been now enforced by the health agencies in public places. This makes the facial based analysis all the more challenging. 

Face occlusion has been effectively addressed in the scope of face identification solutions \cite{opitz2016grid, wang2017face}. Furthermore, obtaining occlusion invariant face recognition frameworks have been a well studied research challenge. However, it is important to note that most of the prior works address generic occlusion that commonly appear in in-the-wild images, such as sunglasses and partial faces. Similar to identity recognition, describing attributes from face images in the wild has also been quite popular \cite{yang2020hierarchical, mao2020deep}. To describe attributes automatically from masked facial images can be very tricky but very advantageous. It can not only be used to build identifiers directly \cite{kumar2011describable}, but also enable construction of scalable hierarchical datasets, further benefiting image classification and attribute-to-image generation tasks \cite{yan2016attribute2image, yan2014hd}. Moreover, attribute prediction can help in tagging vulnerable age groups in a social collaborative environment. 

Traditionally, the generic procedure of predicting face attributes involved constructing low level descriptors through landmark detection and then building domain classifiers for prediction. In this paper, we propose a landmark free end-to-end deep multi-task learning approach for masked facial attribute prediction. The heterogeneous set of attributes include ordinal (age) and nominal (gender and ethnicity) values. Our paper presents three contributions: (1) To the best of our knowledge, we are the first to utilize deep multi-task learning in holistic attribute prediction in masked facial images. (2) We encode heterogeneity and attribute correlation with convolutional neural networks (CNNs) which consists of shared feature learning for attribute prediction, and category-specific feature learning for heterogeneous attributes. (3) We evaluate the proposed approach for attribute prediction to demonstrate superior performance on benchmark face attribute based UTKFace dataset while using only 0.3M parameters in our convolutional encoder. We also provide explainability of the proposed model performance by visualizing the class activation maps on masked facial images. The paper is organized as follows. In Sec. \ref{sec:method}, we discuss the proposed methodology. In Sec. \ref{sec:results}, we give the experimental results and analysis followed by conclusion in Sec. \ref{sec:conclu}.
\begin{figure*}[thpb]
       \centering
       \fbox{
       \includegraphics[scale=0.45]{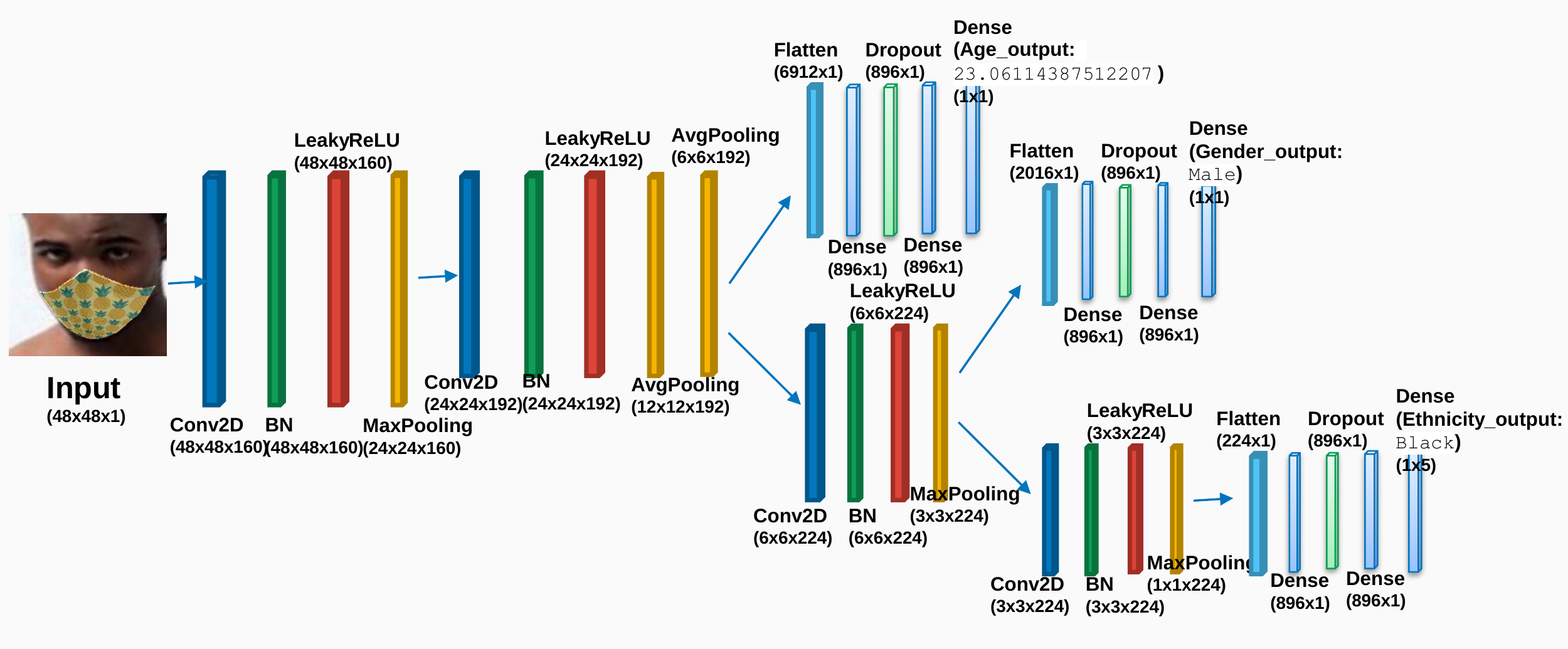}}
       \caption{Proposed \textit{MaskMTL} architecture with hierarchical parameter sharing.}
       \label{fig:method}
\end{figure*}


\section{Proposed Methodology}
\label{sec:method}
We present the proposed MaskMTL architecture with hierarchical parameter sharing as shown in Fig. \ref{fig:method}  \cite{caruana1997multitask, duong2015low, ruder2017overview}. The proposed encoder consists of three convolution layers and is a common module present in all our multi-task learning architectures. In hierarchical parameter sharing, we have different number of layers to extract the features from encoder based on specific tasks. Here, the tasks are divided into main task and auxiliary tasks based on the complexity of inference \cite{Multitaskingformachines}. The auxiliary tasks such as age estimation and gender recognition try to regularize the main task of ethnicity recognition. There are two dense layers in first auxiliary task (age estimation) branch to extract features from encoder. In the second auxiliary task (gender recognition) branch, there is a convolution layer followed by two dense layers to extract the features from encoder. The main task (ethnicity recognition) extract the features from the convolution layer of second auxiliary task through a branch consisting of a convolution layer followed by two dense layers. 
\subsection{Multi-task learning} 
In deep neural networks, multi-task learning can be handled with even a single network that can learn multiple subtasks \cite{caruana1997multitask, hyun2017recognition}. Fig. \ref{fig:method}  illustrates a structure for typical neural networks used for multi-task learning where task generic layers are shared amongst subtasks but task specific layers are separate for every subtask \cite{ruder2017overview}.
\begin{figure}[t]
       \centering
       \fbox{
       \includegraphics[scale=0.48]{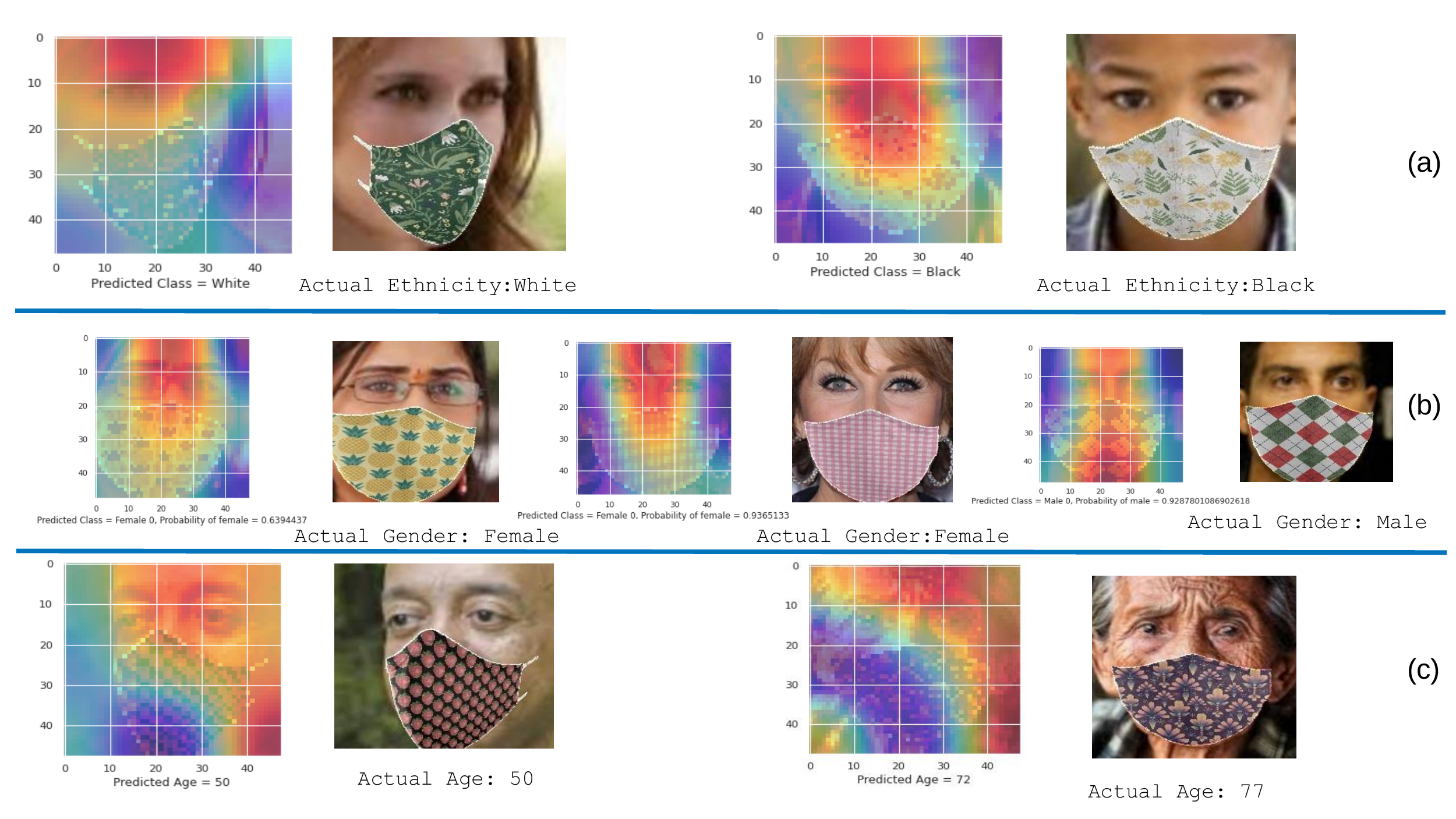}}
       \caption{Class activation map results of a) Ethnicity b) Gender and c) Age. (Red regions denote where maximum activation is attained.) }
       \label{figure:cam}
\end{figure}
In hard parameter sharing of MTL, the hidden layers are shared between all the tasks while keeping several task specific output layers. This reduces the risk of overfitting. Intuition behind it is that the model finds a common representation that captures all the tasks through shared layers and hence there is less chance of overfitting on the original task. While in soft parameter sharing of MTL, each task has its own hidden layers (model) with its own weights. The weights of hidden layers in every tasks are then regularized to overcome overfitting on a particular task.

In the hierarchical parameter sharing, benefits of both hard parameter and soft parameter sharing are leveraged together. The shared hidden layers learn common representation while weights of task specific layers of auxiliary tasks are regularized to improve the performance of the main task.

More precisely, the task generic layers work for every subtask, and only task specific layers work for one of the subtasks. While training, each output layer just predicts the target value that is set for its corresponding subtask, and the weight is updated through gradient back-propagation. In this way, the gradients for multiple subtasks are reflected in the updated weights in terms of improving the performance of all tasks simultaneously. 
\subsubsection{Parameter sharing in neural networks}
The widely used approach for multi-task learning (MTL) with neural networks (NNs) is hard parameter sharing in which a common space representation is learned that generalize well with all subtasks and reduces the risk of overfitting \cite{caruana1997multitask}. The other approach called soft parameter sharing adds a constraint to encourage similarities among loosely coupled space representation of all the tasks. In hierarchical parameter sharing, a regularizing auxiliary task, that takes into account human like context of the main task, is added at different layers to allow the network to learn a semantic space representation at the given layer and further improve the performance of the main task \cite{Multitaskingformachines, zamir2018taskonomy}. In our experiments, we demonstrate that using age estimation and gender recognition as auxiliary tasks helps predicting the ethnicity in non-masked and masked faces.

\begin{figure}[ht]
       \centering
       \fbox{
       \includegraphics[scale=0.11]{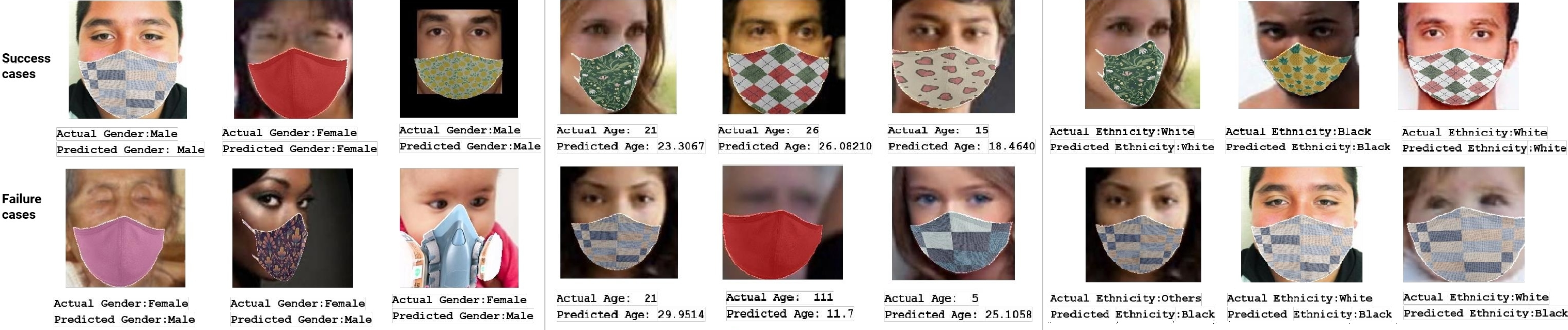}}
       \caption{Qualitative prediction results of a) Age b) Gender and c) Ethnicity. }
       \label{figure:qualitative}
\vspace{-2\baselineskip}
\end{figure}




\subsubsection{Non-local attention}
The facial attribute features learned using neural network in non-masked face data plays significant contribution in age, gender and ethnicity prediction. The facial information is scarcely available in masked face data. Since we generate masked faces by augmenting masks to non-masked faces, the data distribution changes and a new domain is created. Hence we separately train the neural network for masked faces. For masked faces, we introduce non-local attention \cite{wang2018non} after the encoder to focus only on visible facial features and improve the performance of model. Here, the non-local attention operation computes the response at a position as a weighted sum of features at all the positions \cite{buades2005non}.

\subsection{Learning MaskMTL architecture} 


To train our proposed model for age, gender and ethnicity prediction, we select L1-loss, categorical cross entropy (CCE) loss and binary cross entropy (BCE) loss as our objective function and back propagate the corresponding gradients to update the weights for each attribute respectively.

\section{Experimental results and analysis}
\label{sec:results}


\subsection{Dataset}
The two tasks required to create an unbiased masked face dataset includes i) selection of a face dataset that has minimal bias towards attributes to be learnt (i.e. age, gender, race) and ii) have a masking strategy that generates different types (color, texture, size) of masks on varying face poses. The first task is solved by ``The UTKFace Dataset" \cite{zhifei2017cvpr}. It comprises of 20,000+ images in three parts (I, II and III), comprising of 10437, 10719 and 3252 images, respectively with ages varying from 0 to 116 years along with 5 different races (Indian, Asian, White, Black, Others) as well as a label for gender (Male/Female). The dataset has plethora of variance in terms of illumination changes, pose variations, occlusion, spatial resolution and scale. We utilize part I as training split, part II as testing split and part III as validation set for fair comparison.\\
\\\textit{MaskTheFace:} To compute masked images, we utilise MaskTheFace \cite{anwar2020masked}. It generates masks with multiple texture and color variations (N95, cloth, fiber, etc.) with wide face angle coverage. We apply MaskTheFace \cite{anwar2020masked} on the UTKFace dataset to generate a masked UTKFace dataset for our task.
\begin{table}[t]
\small
\centering
\begin{tabular}{ccccccc}
\hline
Method                                                               & Encoder                                                                             & Resolution     & Face Mask    & \begin{tabular}[c]{@{}c@{}}Gender\\ (Accuracy)\end{tabular} & \begin{tabular}[c]{@{}c@{}}Race\\ (Accuracy)\end{tabular} & \begin{tabular}[c]{@{}c@{}}Age\\ (L1-Loss)\end{tabular} \\ \hline
Single Task\cite{lin2019pareto}                                                          & ResNet 18                                                                           & 224x224        & No           & 0.8791                                                    & 0.7812                                                  & 13.55                                                \\
Fixed MTL\cite{lin2019pareto}                                                            & ResNet 18                                                                           & 224x224        & No           & 0.8739                                                    & 0.7838                                                  & 13.90                                                \\
GradNorm\cite{lin2019pareto}                                                             & ResNet 18                                                                           & 224x224        & No           & 0.8809                                                    & 0.7660                                                  & 13.79                                                \\
Uncertainty\cite{lin2019pareto}                                                          & ResNet 18                                                                           & 224x224        & No           & 0.8796                                                    & 0.7704                                                  & 13.96                                                \\
MOO-MTL\cite{lin2019pareto}                                                              & ResNet 18                                                                           & 224x224        & No           & 0.8780                                                    & 0.7784                                                  & 13.80                                                \\
Pareto MTL\cite{lin2019pareto}                                                           & ResNet 18                                                                           & 224x224        & No           & 0.8855                                                    & 0.7928                                                  & 13.51                                                \\

\textbf{Ours}                                                        & \textbf{\begin{tabular}[c]{@{}c@{}}3 Layer Hierarchical\\ Parameter Sharing\\ + Non Local\end{tabular}} & \textbf{48x48} & \textbf{No}  & \textbf{0.8965}                                                & \textbf{0.7852}                                              & \textbf{11.57}                                            \\ \hline
Ours                                                                 & \begin{tabular}[c]{@{}c@{}}3 Layer Hard\\ Parameter Sharing\end{tabular}         & 48x48          & Yes          & 0.8842                                                       & 0.6850                                                     & 14.90                                                    \\
Ours                                                                 & \begin{tabular}[c]{@{}c@{}}3 Layer Soft\\ Parameter Sharing\end{tabular}         & 48x48          & Yes          & 0.8686                                                       & 0.6315                                                     & 15.90                                                   \\
Ours                                                                 & \begin{tabular}[c]{@{}c@{}}3 Layer Hierarchical\\ Parameter Sharing\end{tabular}         & 48x48          & Yes          & 0.8801                                                       & 0.7425                                                     & 12.60                                                    \\ \hline
\textbf{\begin{tabular}[c]{@{}c@{}}MaskMTL \\ (Ours)\end{tabular}} & \textbf{\begin{tabular}[c]{@{}c@{}}3 Layer Hierarchical\\ Parameter Sharing\\ + Non Local\end{tabular}} & \textbf{48x48} & \textbf{Yes} & \textbf{0.8913}                                              & \textbf{0.7927}                                            & \textbf{12.15}                                          \\ \hline
\end{tabular}
\caption{ The gender accuracy, race accuracy, and age L1-loss obtained by different algorithms. Various ablation studies are performed with the proposed method on diverse masked faces with comparable accuracy.}
\vspace{-2\baselineskip}
\label{tab:tab1}
\end{table}

\textit{Implementation details:}
The library is written in python using Tensorflow's Keras framework and trained using single Nvidia K80 GPU (Google Colab). The model was trained in an end-to-end manner using proposed multi-task loss. The proposed network is trained from scratch using Adam optimizer with 32 batch size and 0.001 learning rate for 40 epochs. We train our model on both non-masked and masked UTKFace dataset resized to 48x48 resolution with same training hyper-parameters for fair evaluation.
\subsection{Quantitative Results}
Our encoder has only 0.3M parameters when compared to 11M parameters in ResNet-18, but performs better than other multi-task methods while utilising lower resolution images for predicting age, gender and ethnicity. 
Table \ref{tab:tab1} compares our models trained on both mask/no-mask datasets. Our model performs drastically well compared to other state-of-the-art non-masked models that use mechanisms such as GradNorm, Uncertainity etc. \cite{lin2019pareto} for learning hyper-parameters used in the same multi-task loss. Without using any weighted multi-task loss, hierarchical parameter sharing ensures that the features learnt by one task are used for solving the subsequent task learns optimal features at all levels. Thus, leads to improvements in classification in age, gender and ethnicity. Also, unlike ResNet18, our model is trained from scratch and shall improve further if pretrained using some other face dataset.

\begin{table}[ht]
\centering
\begin{tabular}{ccccccccccc}
                       &                              &                                                   &  &                        & \textit{W}                            & \textit{B}                           & \textit{A}                           & \textit{I}                           & \textit{O}                                               &  \\ \cline{6-10}
                       &                              &                                                   &  & \multicolumn{1}{c|}{\textit{W}} & \cellcolor[HTML]{009901}1651 & 292                         & 106                         & 199                         & \multicolumn{1}{c|}{30}                         &  \\
                       & \textit{M}                            & \textit{F}                                                 &  & \multicolumn{1}{c|}{\textit{B}} & 31                           & \cellcolor[HTML]{32CB00}924 & 17                          & 49                          & \multicolumn{1}{c|}{3}                          &  \\ \cline{2-3}
\multicolumn{1}{c|}{\textit{M}} & \cellcolor[HTML]{F8FF00}2017 & \multicolumn{1}{c|}{112}                          &  & \multicolumn{1}{c|}{\textit{A}} & 27                           & 44                          & \cellcolor[HTML]{34FF34}628 & 30                          & \multicolumn{1}{c|}{8}                          &  \\
\multicolumn{1}{c|}{\textit{F}} & 633                          & \multicolumn{1}{c|}{\cellcolor[HTML]{FCFF2F}1903} &  & \multicolumn{1}{c|}{\textit{I}} & 52                           & 134                         & 17                          & \cellcolor[HTML]{67FD9A}695 & \multicolumn{1}{c|}{7}                          &  \\ \cline{2-3}
                       &                              &                                                   &  & \multicolumn{1}{c|}{\textit{O}} & 80                           & 72                          & 33                          & 83                          & \multicolumn{1}{c|}{\cellcolor[HTML]{9AFF99}93} &  \\ \cline{6-10}
                       &                              &                                                   &  &                        &                              &                             &                             &                             &                                                 & 
\end{tabular}
\caption{Confusion Matrix for a) Gender classification (\textit{M}ale and \textit{F}emale) and b) Ethnicity classification (\textit{W}hite, \textit{B}lack, \textit{A}sian, \textit{I}ndian, \textit{O}thers.}
\label{tab:tab2}
\end{table}
\vspace{-2\baselineskip}

Table \ref{tab:tab1} also compares the results of our mask model with our no-mask model across different parameter sharing architectures. Although we notice slight decrease in accuracy, the results are comparable given that the model has less facial features to learn. The hierarchical parameter sharing architecture performs the best when compared to standard hard and soft parameter sharing models as shown in Table \ref{tab:tab1}. We tried to finetune our no-mask trained model for our masked dataset but it didn't converge. The results also validate the inclusion of non-local attention block in network that further increases prediction accuracy by considerable amount. 

\begin{table}[ht]
\centering
\begin{tabular}{l|c|c|c} \hline
\textbf{Class}  & \textbf{Precision} & \textbf{Recall} & \textbf{F1-Score} \\ \hline
\textbf{White}  & 0.90               & 0.72            & 0.80              \\
\textbf{Black}  & 0.63               & 0.90            & 0.74              \\
\textbf{Asian}  & 0.78               & 0.85            & 0.82              \\
\textbf{Indian} & 0.66               & 0.77            & 0.71              \\
\textbf{Others} & 0.66               & 0.26            & 0.37              \\
\textbf{Male}   & 0.81               & 0.96            & 0.88              \\
\textbf{Female} & 0.94               & 0.75            & 0.84    \\ \hline         
\end{tabular}
\caption{Precision, Recall and F1-Score for all classes trained on masked dataset.}
\label{tab:tab3}
\end{table}
\vspace{-2\baselineskip}

Table \ref{tab:tab2} and \ref{tab:tab3} describe the classification results with precision and recall values for all classes. The model is able to correctly predict all classes. The greatest challenge visible from these results is the low recall of ``Others" ethnicity class which can be due to presence of similar attributes in other ethnicity.

\vspace{-0.18in}
\subsection{Qualitative Results}
In Fig. \ref{figure:qualitative}, we provide some examples of the success and failure cases on the UTKFace dataset. We observe that in failure cases are observed in attribute age when the image is blurred or facial features are quite smooth (as in case of child (actual age:5) and young adolescent (predicted age: 25). In gender, the failure cases are observed in side profiles or baby where there are no hairs on head and distinguishable features. In case of ethnicity prediction, model gets confused when there is similarity between the races or the contrast of the image is not well adjusted. In Fig. \ref{figure:cam}, we provide explanability to the prediction models by providing the class activation maps. 
In case of ethnicity, we observe that the attention is focused on forehead regions where there is a major patch of the skin tone. In case of gender, the attention is on the forehead bindi (Race: Asian-Indian) or bangs (in hair) for the female whereas for male it is the nose and forehead regions. In case of age, we observe that it is the wrinkles which gets highlighted in the activation maps which indeed is the determining factor of attribute age.
\vspace{-0.15in}
\section{Conclusion}
\label{sec:conclu}
We proposed an end-to-end multi-task learning based attribute prediction framework for ethnicity, gender and age classification while formulating a hierarchical parameter sharing strategy that learns the same on the UTKFace dataset. We utilised a masking approach to create masked UTKFace dataset. Our lightweight model achieved comparable accuracy even on masked images while surpassing ParetoMTL in standard UTKFace dataset. In future, we intend to explore our work for learning other facial attributes.

\bibliographystyle{splncs04}
\bibliography{refs}

\end{document}